%% file: iclr2019_conference.tex
\documentclass{article} 
\usepackage{iclr2019_conference,times}

\input{math_commands.tex}

\usepackage{hyperref}
\usepackage{url}
\usepackage{graphicx}

\definecolor{temp}{HTML}{FF0000}
\definecolor{todo}{HTML}{FFA419}
\definecolor{done}{HTML}{0BF9AB}
\definecolor{info}{HTML}{1E90FF} 
\definecolor{moresamples}{HTML}{00FF00}

\title{Can Graph Neural Networks Go ``Online''? An Analysis of Pretraining and Inference}

\author{Lukas Galke\\
  Kiel University\\
  Germany\\ 
  \texttt{lga@informatik.uni-kiel.de}\\
  \And
  Iacopo Vagliano\\
  ZBW -- Leibniz Information Centre for Economics\\
  Kiel, Germany\\
  \texttt{I.Vagliano@zbw.eu}\\
  \And
  Ansgar Scherp\\
  University of Essex\\
  United Kingdom\\
  \texttt{as18255@essex.ac.uk}
}

\iclrfinalcopy 

\begin{document}

\maketitle

\begin{abstract}
  Large-scale graph data in real-world applications is often not static but dynamic, i.\,e., new nodes and edges appear over time.
  Current graph convolution approaches are promising, especially, when all the graph's nodes and edges
  are available during training.
  When unseen nodes and edges are inserted after training, it is not yet
  evaluated whether up-training or re-training from scratch is preferable.
  We construct an experimental setup, in which we insert previously unseen
  nodes and edges after training and conduct a limited amount of inference epochs. 
  In this setup, we compare adapting pretrained graph neural networks against retraining from
  scratch.
  Our results show that pretrained models yield high accuracy scores on the
  unseen nodes and that pretraining is preferable over retraining from scratch. 
  Our experiments represent a first step to evaluate and develop
  truly online variants of graph neural networks.
\end{abstract}
\section{Introduction}
Recent advances in graph neural networks yield promising results.
Various methods have recently emerged for applying neural networks to
graph-structured
data~\citep{DBLP:journals/corr/BrunaZSL13,DBLP:journals/corr/HenaffBL15,NIPS2015_5954,DBLP:journals/corr/LiTBZ15,DBLP:conf/nips/DefferrardBV16}.
%
Notably, graph convolutional
networks~\citep{DBLP:journals/corr/KipfW16a,DBLP:journals/corr/KipfW16}
along with their extensions such as graph attention
networks~\citep{velickovic2018graph} have achieved promising results.
When evaluating these approaches, a typical assumption is that the whole graph
structure (nodes and edges) is fully known during training.
It is yet less explored whether continuing the training process on
unseen data is a valid approach for dealing with updates in the graph structure.

This aspect is important for large-scale, dynamic graphs such as social networks, citation networks, or in other online applications, where new nodes and edges appear over time~\citep{Aggarwal:2014:ENA:2620784.2601412}.
For instance, consider new users or new items in a recommendation setup or newly published papers in a classification setup.
Re-training a large-scale model from scratch whenever new nodes or edges appear might be costly, and thus, undesirable.
We identify the lack of inference capabilities as a potential drawback of graph convolution and other recent approaches on graph data.


Tasks in which the full graph structure is known during training are known as
\textit{transductive} settings~\citep{DBLP:journals/corr/KipfW16}. 
Settings with unseen structure are called
\textit{inductive}~\citep{DBLP:conf/icml/YangCS16,DBLP:conf/nips/HamiltonYL17,velickovic2018graph}.
The approaches are, however, oftentimes categorized as being either suited for
inductive learning or not~\citep{DBLP:conf/icml/YangCS16}. The possibility to
conduct few inference epochs with a pre-trained model is oftentimes discussed
yet rarely evaluated.

In this paper, we create a dedicated experiment to evaluate inference capabilities of graph neural networks.
We apply the networks on two train-test settings, one with many labelled nodes and one with few labelled nodes. 
After training, unseen nodes and edges are
inserted into the graph and the models may perform a limited
amount of parameter updates. We closely analyze the test accuracy after each
of these inference epochs, while comparing the networks performance with pretraining versus without pretraining.
%
The models considered in our experiments include graph convolutional
networks~\citep{DBLP:journals/corr/KipfW16}, graph attention
networks~\citep{velickovic2018graph},
and GraphSAGE~\citep{DBLP:conf/nips/HamiltonYL17} on three datasets (Cora, Citeseer, Pubmed), which we cast into two complementary inductive settings.
In total, we have conducted 60 experiments with 100 repetitions each to alleviate random effects.
Our findings suggest that pretraining is indeed beneficial for all considered models.
This result holds for both of the train-test settings, with many labelled nodes and with few labelled nodes. 

In summary, our contributions are two-fold: (1)~We provide an
experimental setup to evaluate inference capabilities of graph neural networks.
(2)~We offer empirical evidence that pretraining is useful in both cases: when the
train-test ratio is high and when it is low. 

We describe our experimental setup in
\Secref{sec:setup} and the applied methods in \Secref{sec:methods}.
We present the results in \Secref{sec:results}, which we
discuss in \Secref{sec:discussion}, before we conclude.

\section{Experimental Setup}\label{sec:setup}

\begin{table}[t]
  \caption{Statistics for train-test splits: few-many (A) and many-few (B) settings on the citation networks
    datasets: Cora, Citeseer, and Pubmed. The unseen nodes and edges are available
    only after the training epochs. The test samples for measuring accuracy are
  a subset of the unseen nodes. The label rate is the percentage of labelled nodes for training.}
  \label{tab:datasets}
  \begin{center}
    \begin{tabular}{lrrrrrr}
      \textbf{Dataset} & \multicolumn{2}{r}{\bf Cora}  &\multicolumn{2}{r}{\bf Citeseer} & \multicolumn{2}{r}{\bf Pubmed} \\\hline
      Classes & \multicolumn{2}{r}{7} & \multicolumn{2}{r}{6} & \multicolumn{2}{r}{3}\\
      Features & \multicolumn{2}{r}{1,433} &\multicolumn{2}{r}{3,703} & \multicolumn{2}{r}{500}\\
      Nodes & \multicolumn{2}{r}{2,708} & \multicolumn{2}{r}{3,327} & \multicolumn{2}{r}{19,717}\\
      Edges & \multicolumn{2}{r}{5,278} & \multicolumn{2}{r}{4,552} & \multicolumn{2}{r}{44,324}\\
      Avg. Degree & \multicolumn{2}{r}{3.90} & \multicolumn{2}{r}{2.77} &
      \multicolumn{2}{r}{4.50}\\\hline\\
      \textbf{Setting} & \textbf{A} & \textbf{B} & \textbf{A} & \textbf{B} & \textbf{A} & \textbf{B} \\ \hline 
      Train Samples  & 440        & 2,268      & 620        & 2,707      & 560        & 19,157     \\
      Train Edges    & 342        & 3,582      & 139        & 2,939      & 34         & 41,858        \\
      Unseen Nodes   & 2,268      & 440        & 2,707      & 620        & 19,157     & 560     \\
      Unseen Edges   & 4,936      & 1,696      & 4,413      & 1,613      & 44,290     & 2,466     \\
      Test Samples   & 1,000      & 440        & 1,000      & 620        & 1,000      & 560        \\
      Label Rate     & 16.2\%      & 83.8\%      & 18.6\%      & 81.4\%      & 2.8\% & 97.2\%      \\\hline
    \end{tabular}
  \end{center}
\end{table}

We construct a dedicated experimental setup to evaluate the inference
capabilities of graph neural networks. We include edges in the training set if
and only if both its source and destination node are both in the training set.
The training process is then split in two steps. First, we pre-train the model on
the labelled training set. Then, we insert the previously unseen
nodes and edges into the graph and continue training for a limited amount of
inference epochs. The unseen nodes do not introduce any new labels. Instead, the
unseen nodes provide features and may be connected to known labelled nodes. We
evaluate the accuracy on the test nodes, which are a subset of the unseen nodes,
before the first and after each inference epoch. For each model, we compare
using 200 pretraining epochs versus no pretraining. In the latter case, the
training begins during inference, which is equivalent to retraining from
scratch whenever new nodes and edges are inserted. This allows us to assess
whether pretraining is helpful for applying graph neural networks on dynamic
graphs.

\paragraph{Datasets} We impose our experimental setup on the three standard
citation datasets: Cora, Citeseer, and
Pubmed~\citep{DBLP:journals/aim/SenNBGGE08}.
Nodes are research papers represented by textual features and
annotated with a class label. Edges resemble citation relationships.
These datasets are often used in transductive
setups~\citep{DBLP:conf/icml/YangCS16,DBLP:journals/corr/KipfW16,velickovic2018graph}.
In our experimental setup with unseen nodes, however, we cast these datasets to be
inductive. The citation edges are regarded as undirected connections and we add
self-loops~\citep{DBLP:journals/corr/KipfW16}.

\paragraph{Few-many setup (A) and many-few setup (B)}
We use two different train-test splits for each dataset.
Setting A is derived from the train-test split for transductive
tasks~\cite{DBLP:journals/corr/KipfW16}. It consists of few labeled nodes that
induce our training set and many unlabeled nodes.
Setting B instead comprises many training nodes and few test nodes. We set it up
by inverting the train-test mask of Setting A and assign the edges accordingly.
Setting B is motivated from applications, in which a large graph is already known and incremental changes occur over time, such as for citation recommendations, link prediction in social networks, and others~\citep{Aggarwal:2014:ENA:2620784.2601412,Galke:2018:MAA:3209219.3209236}. 
We refer to Table~\ref{tab:datasets} for the details of the datasets and the two
settings.

\section{Models}\label{sec:methods} 
We compare the
inference capabilities of graph convolutional networks
(GCN)~\citep{DBLP:journals/corr/KipfW16}, GraphSAGE~\citep{DBLP:conf/nips/HamiltonYL17}, and  graph attention networks
(GAT)~\citep{velickovic2018graph},
The hidden representation of each node $i$ in layer $l$ is defined as:
\begin{equation*}
h_i^{(l+1)} = \sigma \left( \sum_{j \in \mathcal{N}(i)} \frac{1}{c_{ij}} W^{(l)} h_j^{(l)} \right)
\end{equation*}
where $\mathcal{N}(\cdot)$ refers to the set of adjacent nodes and $\sigma$ is a
nonlinear activation function. The normalizing factor $c_{ij}$ depends on the
respective model: GCN use $c_{ij} = \sqrt{|\mathcal{N}(i)|} \cdot
\sqrt{|\mathcal{N}(j)|}$, GraphSAGE with mean aggregation uses $c_{ij} =
|\mathcal{N}(i)|$,
and GAT use learned attention weights instead of $\frac{1}{c_{ij}}$ which are
computed by a nonlinear transformation from the concatenation of $h_i^{(l)}$ and
$h_j^{(l)}$.
All employed graph neural networks use two graph convolution layers that aggregate neighbor
representations.  The output dimension of the second layer corresponds to the number of classes.
Thus, the features within the two-hop neighborhood of each labelled node are
taken into account for its prediction.

\paragraph{Hyperparameters}
We adopt the same hyperparameter values as proposed in the original
works. For GCN, we use 16 or 64 (denoted by GCN-64) hidden units per layer, ReLU
activation, 0.5 dropout rate, along with an (initial) learning rate
of 0.005 and weight decay $5 \cdot
10^{-4}$~\citep{DBLP:journals/corr/KipfW16}.
For GAT, we use 8 hidden units per layer and 8
attention heads on the first layer. The second layer has 1 attention head (8 on
Pubmed). We set the learning rate  to 0.005 (0.01 on Pubmed) with weight decay
0.0005 (0.001 on Pubmed)~\citep{velickovic2018graph}. 
For GraphSAGE, we use 64 hidden units per layer with mean aggregation, ReLU
activation, and a dropout rate of 0.5. 
We set the learning rate to 0.01 with weight decay $5 \cdot
10^{-4}$~\citep{DBLP:conf/nips/HamiltonYL17}.
Our MLP baseline has one hidden layer with 64 hidden units, ReLU
activation, a dropout rate of 0.5, learning rate 0.005 and weight decay $5 \cdot
10^{-4}$.
In all cases, we initialize weights according to
\citet{DBLP:journals/jmlr/GlorotB10} and use Adam~\citep{DBLP:journals/corr/KingmaB14} to optimize cross-entropy.
We use a fresh initialization of the optimizer at the beginning of the inference epochs.
We do not use any early stopping in our experiments.
%
%

\section{Results}\label{sec:results}
\begin{figure}[ht] 
  \centering
  \includegraphics[width=\textwidth]{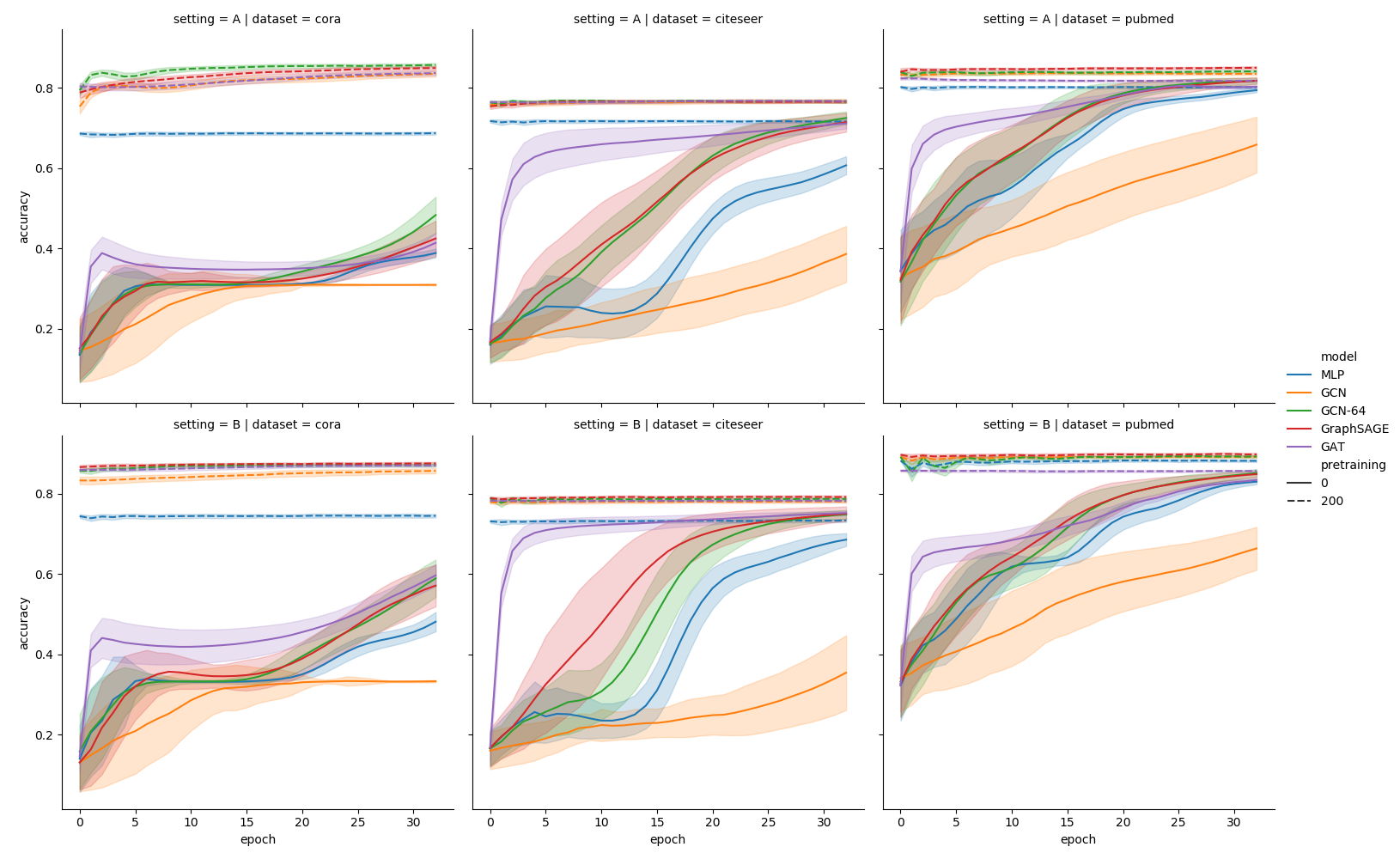}
  \caption{Test accuracy after each inference epoch for the many-few settings A \figtop{}
    and few-many setting B \figbottom{} on the datasets Cora, Citeseer, and Pubmed. Each line resembles the mean
    of 100~runs and its region shows the standard deviation.
    The dashed lines show the results with 200~pretraining. The solid lines are the results without prertraining.
  }\label{fig:results}
\end{figure}

\Figref{fig:results} shows the results of the three models on the
three datasets: Cora, Citeseer, and Pubmed. Pretrained models score consistently
higher than non-pretrained models while having substantially less variance.
The accuracy of the pretrained models plateaus after few inference epochs (up to 10 on
Cora-A and Pubmed-B).
Without any pretraining, GAT shows the fastest learning process. 
The absolute scores of pretrained graph neural networks are higher than the
ones of MLP. From a broad perspective, the scores of pretrained graph neural
networks are all on the same level. While GCN falls behind the others on Cora-B,
GAT falls behind the others on Pubmed.

The absolute scores of the many-few setting B are higher than the ones of few-many setting A by a
constant margin. We globally compare the results of setting A and B by measuring the
Jensen-Shannon divergence~\citep{DBLP:journals/tit/Lin91} between the accuracy distributions.
The Jenson-Shannon divergence between the two settings is lower with
pretraining (between 0.0057 for GAT and 0.0115 for MLP) than it is without
pretraining (between 0.0666 for GraphSAGE to 0.1013 for GCN).


\section{Discussion}\label{sec:discussion}
We have developed an experimental setup to evaluate inferencing
capabilities of graph neural networks, which we use to conduct inductive
experiments on the well-known citation graph datasets: Cora, Citeseer, and Pubmed.
Our results show that graph neural networks perform well even though we insert new
nodes and edges after training.
For the three datasets considered in this study, the accuracy plateaus after
very few inference epochs. 
This observation holds for both train-test split settings: many-few and few-many.
We verified that the accuracy distributions are similar in both train-test
splits by measuring the Jenson-Shannon divergence.
The low variances of the pretrained models indicate that the 100 runs for each of the pretrained models converge to
yield similar accuracy scores and that they are robust against the addition of unseen nodes.

The employed models all consist of two layers. Thus, the models only exploit
node features in the two-hop neighborhood of each labeled node.
Technically, also more distant nodes' features would be available, especially
during inference. Model depth is, however, still an open
issue~\citep{DBLP:journals/corr/KipfW16} in the graph domain. Deeper models
could, in theory, exploit more distant nodes' features.

Evaluating inferencing capabilities is important because full
re-training might not be feasible on large graphs.
We have taken a first step to bring current research in graph neural
networks closer to practical applications.
Our setup is close to real-world
applications, where new nodes appear dynamically over time.
Our findings show that maintaining one model and continuing the
training process when the data changes is a valid approach for dealing with dynamic
graphs.

\section{Conclusion}

Pretrained graph neural networks yield high accuracy with low variance even
though new nodes and edges are inserted into the graph.
This property is mandatory for applying graph neural networks to large-scale,
dynamic graphs as often found in real-world scenarios.

\subsubsection*{Reproducibility and reusablity}
The source code to reproduce our experiments, to impose our experimental setup on
other datasets, or to include further models, is available under
\href{https://github.com/lgalke/gnn-pretraining-evaluation}{\url{github.com/lgalke/gnn-pretraining-evaluation}}.


\bibliography{iclr2019_conference}
\bibliographystyle{iclr2019_conference}


%

\end{document}

%% file: math_commands.tex
\usepackage{amsmath,amsfonts,bm}

\newcommand{\figtop}{{\em (Top)}}
\newcommand{\figbottom}{{\em (Bottom)}}

\def\Figref#1{Figure~\ref{#1}}

\def\Secref#1{Section~\ref{#1}}

\def\eqref#1{equation~\ref{#1}}

\def\1{\bm{1}}

\DeclareMathAlphabet{\mathsfit}{\encodingdefault}{\sfdefault}{m}{sl}
\SetMathAlphabet{\mathsfit}{bold}{\encodingdefault}{\sfdefault}{bx}{n}

